\documentclass[letterpaper, preprint, paper,11pt]{AAS}	
\usepackage{multirow}
\usepackage{graphicx}
\usepackage{booktabs}
\usepackage{array}
\usepackage{bm}
\usepackage{amsmath}
\usepackage[colorlinks=true, pdfstartview=FitV, linkcolor=black, citecolor= black, urlcolor= black]{hyperref}
\usepackage{overcite}
\usepackage{footnpag}			      	
\usepackage{amssymb,amsmath}
\usepackage{booktabs}
\usepackage{caption}
\usepackage{subcaption}
\usepackage{soul}
\usepackage{xcolor}

\usepackage[normalem]{ulem}
\usepackage{mathtools}

\PaperNumber{25-705}

\begin{document}

\title{Understanding and Utilizing Dynamic Coupling in Free-Floating Space Manipulators for On-Orbit Servicing}

\author{Gargi Das\thanks{Postdoctoral Researcher, Department of Aerospace Engineering \& Engineering Mechanics, University of Cincinnati, Cincinnati, OH 45221, USA.},  
Daegyun Choi\thanks{Postdoctoral Researcher, Department of Aerospace Engineering \& Engineering Mechanics, University of Cincinnati, Cincinnati, OH 45221, USA.},
\ and Donghoon Kim\thanks{Assistant Professor, Department of Aerospace Engineering \& Engineering Mechanics, University of Cincinnati, Cincinnati, OH 45221, USA.}
}

\maketitle{}

\begin{abstract}
This study proposes a dynamic coupling-informed trajectory optimization algorithm for free-floating space manipulator systems (SMS{s}). Dynamic coupling between the base and the manipulator arms plays a critical role in influencing the system's behavior. While prior research has predominantly focused on minimizing this coupling, often overlooking its potential advantages, this work investigates how dynamic coupling can instead be leveraged to improve trajectory planning. Singular value decomposition 
of the dynamic coupling matrix is employed to identify the dominant components governing coupling behavior. A quantitative metric is then formulated to characterize the strength and directionality of the coupling and is incorporated into a trajectory optimization framework. To assess the feasibility of the optimized trajectory, a sliding mode control
-based tracking controller is designed to generate the required joint torque inputs. Simulation results demonstrate that explicitly accounting for dynamic coupling in trajectory planning enables more informed and potentially more efficient operation, offering new directions for the control of free-floating SMS{s}. 
\end{abstract}

\section{Introduction}
\subsection{Background}
Since the launch of Sputnik 1 in 1957, over 20,000 spacecraft have been sent into orbit, of which only 11,100 remain operational \cite{ESA2024}. The vast number of defunct spacecraft, resulting from mission failures, has not only incurred billions of dollars in losses but has also contributed significantly to the proliferation of space debris. To mitigate this issue, on-orbit servicing (OOS) using space manipulator systems (SMSs) has been extensively researched over the past few decades \cite{xu2011survey}. OOS refers to space operations that involve inspecting, repairing, refueling, upgrading, and/or repositioning spacecraft to extend their lifespan and enhance their functionality. Notable examples include the Engineering Test Satellite No. 7 (ETS-VII) \cite{yoshida2001zero}, developed by the National Space Development Agency of Japan, and the Orbital Express program, managed by the United States Defense Advanced Research Projects Agency \cite{ogilvie2008autonomous}, both of which successfully demonstrated robotic servicing capabilities in space. 

SMSs are generally classified into free-flying and free-floating systems based on how the base spacecraft is managed during manipulation tasks \cite{wilde2018equations}. In free-flying systems, the base spacecraft is actively controlled in both translation and rotation using actuators such as thrusters or momentum control devices. In contrast, free-floating systems operate under the conservation of linear and angular momentum, where manipulator motion induces reactive base motion due to the absence of external forces and torques. This intrinsic behavior is known as dynamic coupling.

Dynamic coupling poses both challenges and opportunities. It directly affects the spacecraft's attitude and position, which in turn influences the accuracy of the manipulator's end-effector. Traditionally, base disturbances have been viewed as undesirable, prompting the use of energy-intensive control schemes to suppress them \cite{caccavale2001kinematic,huang2005dynamic,zhang2021attitude}. As a result, free-flying systems have often been the preferred choice for OOS, since their nearly fixed-base condition resembles ground-based manipulators and allows for more straightforward control. However, this approach becomes problematic in close-proximity operations, where thruster firings may inadvertently induce collision risks, contamination, or damage 
nearby satellites 
\cite{soares2002international,rutkovskii2013some}. Furthermore, continuous actuator use can rapidly deplete finite resources like fuel or reaction wheel capacity. Actuator failure can also unintentionally convert a free-flying system into a free-floating one, demanding alternative control strategies.

While free-floating systems offer unique benefits, they also come with challenges, such as reduced end-effector accuracy and complex motion planning due to base disturbances \cite{SEDDAOUI2021311}. Nonetheless, under specific conditions, they present distinct advantages. These systems inherently conserve momentum and do not require active base control, making them ideal for delicate, low-energy, and safe proximity operations. By leveraging the dynamic coupling effect rather than opposing it, free-floating SMSs can perform precise manipulations while preserving system integrity and resource efficiency. The concept of the dynamic coupling factor quantifies the extent of base disturbances caused by manipulator motion \cite{xu1993measure}. Over the past three decades, extensive research has been conducted to understand and mitigate dynamic coupling effects. This includes the formulation of the generalized Jacobian matrix 
\cite{umetani1989resolved}, the introduction of virtual manipulator representations \cite{vafa1990kinematics,liang1997dynamically,yan2014base,xu2016hybrid, peng2020modeling}, the development of dynamic coupling models for a multi-arm SMS \cite{zhou2019dynamic}, and various control strategies to counteract base disturbances \cite{caccavale2001kinematic,huang2005dynamic,zhang2021attitude}. However, only a limited number of studies \cite{nakamura1990nonholonomic} have attempted to utilize dynamic coupling proactively to optimize control performance and energy efficiency.
\subsection{Problem Statement}
While prior studies have examined how static variables, such as mass distribution, link lengths, and initial configurations, affect dynamic coupling, the role of dynamic variable changes (e.g., joint torque profiles, motion trajectories) remains underexplored. Moreover, most existing studies focus on minimizing dynamic coupling without assessing whether it can instead be exploited to support the operation of the free-floating SMS.

This study addresses that gap by investigating how 
dynamic coupling in a free-floating SMS can be leveraged for optimal trajectory design. 
A singular value decomposition (SVD) of the dynamic coupling matrix is performed to identify and analyze the key components of the coupling behavior, and a formulation is proposed to quantify the dynamic coupling. This information is then incorporated into a dynamic coupling-informed trajectory optimization framework. To validate the feasibility of the obtained trajectory, a sliding mode control (SMC)-based tracking controller is designed, which generates joint torque inputs to track the optimized trajectory. The key contributions of this study include:

\begin{itemize}
    \item The dynamic coupling behavior is analyzed and quantified using two metrics: the normalized Shannon entropy ($H_{norm}$) of the singular value spectrum of the dynamic coupling matrix {in Eq. \eqref{eqn_SVD_2}} and the coupling assistance metric ($\cos(\theta_{a})$) in Eq. \eqref{eqn_SVD_4}. 
    \item An optimal trajectory design algorithm is proposed for the free-floating SMS, leveraging dynamic coupling information based on $H_{norm}$ and $\cos(\theta_{a})$.
     \item An SMC is designed to track the optimized trajectory using joint torques, with no control applied to the satellite base. 
\end{itemize}
The overarching goal is to explore how dynamic coupling can be utilized to develop optimized trajectories and integrate them into control frameworks. This establishes a foundation for intelligent, energy-aware, and robust control architectures tailored for future {OOS} missions, particularly in scenarios demanding high system agility and resilience in uncertain or failure-prone environments.
\section{Free Floating Space Manipulator System}

\subsection{Dynamics and Kinematics of Free-Floating SMS}
\begin{figure}[htbp]
	\centering\includegraphics[width=5.5in]{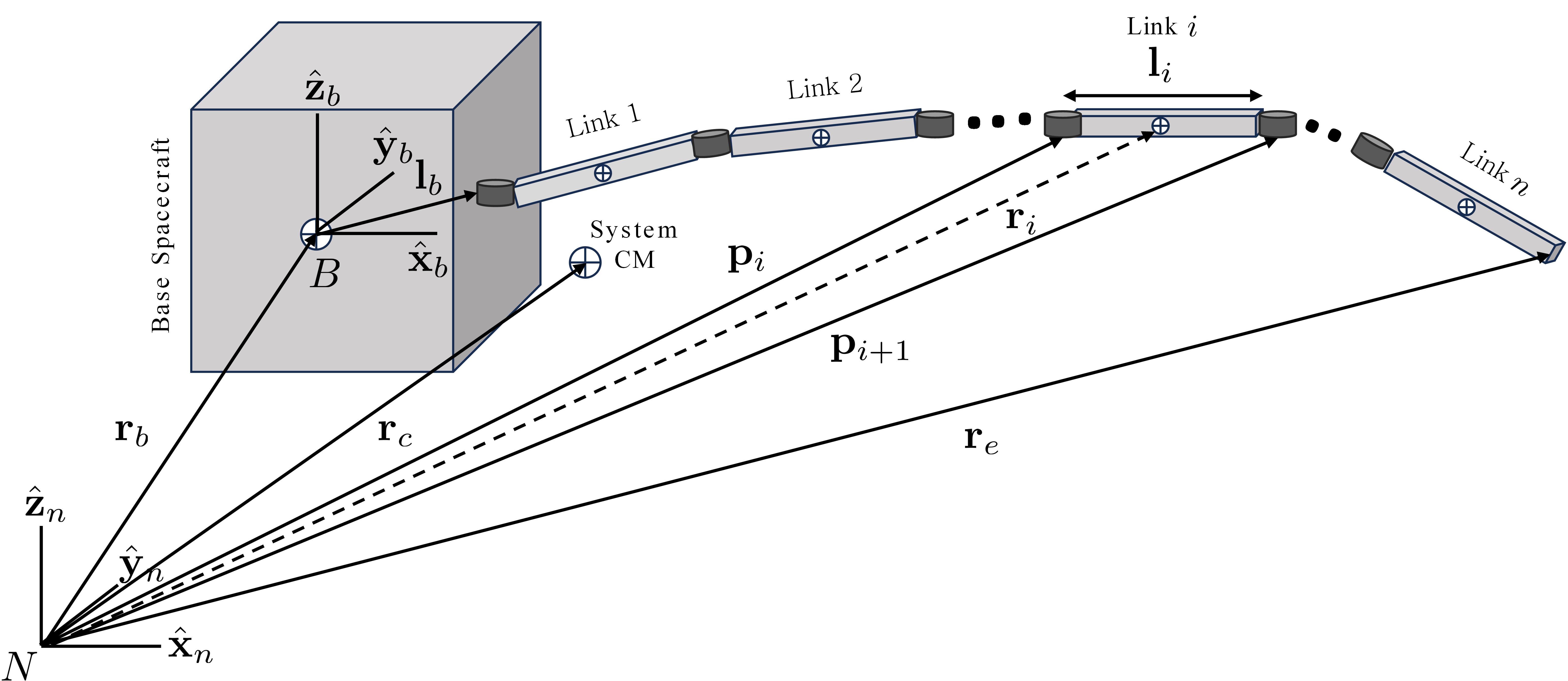}
	\caption{Schematic Diagram of a Free-Floating SMS}
	\label{fig:SMS_diagram}
\end{figure}

Figure \ref{fig:SMS_diagram} shows a free-floating SMS, where the spacecraft base is mounted with an $n$-degree of freedom 
manipulator. 
Euler angles may suffer from a representation singularity; therefore, quaternions are considered to describe the attitude of the base spacecraft with respect to the inertial frame ($N$). The differential kinematic equation of the base spacecraft is defined as \cite{schaub2003analytical}:
\begin{equation}
    \dot{\bm{\epsilon}} = \frac{1}{2} G(\bm{\epsilon}) \boldsymbol{\omega}_b,
\end{equation}
where $\bm{\epsilon}=[\epsilon_1, 
\epsilon_2, 
\epsilon_3, 
\epsilon_4]^T \in \Re^{4}$ is the quaternions, and $\boldsymbol{\omega}_b\in\Re^3$ is the angular velocity of the base spacecraft with respect to the inertial frame. The first three elements ($\epsilon_1, \epsilon_2, \text{ and }\epsilon_3$) form the vector part $\bm{\epsilon}_{\upsilon} =[\epsilon_1,\epsilon_2,\epsilon_3]^T$, and $\epsilon_4$ is the scalar component. 
Note that the quaternions must satisfy the holonomic constraint
: $\bm\epsilon^T\bm\epsilon =1$. The matrix $G(\bm{\epsilon}) \in \Re^{4\times3}$ is defined as: 
\begin{equation}
   G(\bm{\epsilon}) = \begin{bmatrix}
       \epsilon_4 & -\epsilon_3 & \epsilon_2\\
       \epsilon_3 & \epsilon_4 & -\epsilon_1\\
       -\epsilon_2 & \epsilon_1 & \epsilon_4\\
       -\epsilon_1 & -\epsilon_2 & -\epsilon_3
   \end{bmatrix}.
\end{equation}

For the derivation of the dynamics, a rigid space manipulator system is assumed. The end-effector's linear and angular velocities in the inertial frame ($\dot{\bf r}_e\in \Re^{3}$ and $\bm{\omega}_e\in \Re^{3}$) are given by \cite{wilde2018equations}:
\begin{equation}
    \begin{bmatrix}
\dot{\mathbf{r}}_e \\
\boldsymbol{\omega}_e 
\end{bmatrix} = \begin{bmatrix}
    \mathbb{I}_3 & \left[\mathbf{r}_{eb}^{\times}\right]^T\\
    {\mathbb O}_3 & \mathbb{I}_3
\end{bmatrix} \begin{bmatrix}
    \dot{\mathbf{r}}_b\\
  \boldsymbol{\omega}_b
\end{bmatrix} + \begin{bmatrix}
    J_{Te}\\
    J_{Re}
\end{bmatrix} \dot{\mathbf{q}},
\label{Eq_d1}
\end{equation}
where $\dot{\mathbf{r}}_b \in \Re^{3}$ is the linear velocity of the base spacecraft's center of mass, $\mathbb{I}_3\in \Re^{3\times 3}$ is the identity matrix, {and} $\mathbb{O}_3 \in \Re^{3\times 3}$ is the zero matrix. Here, 
$\mathbf{r}_{eb} = \mathbf{r}_e - \mathbf{r}_b$, and $\dot{\mathbf{q}} =\left[\dot{q}_1, \dot{q}_2,...,\dot{q}_n \right]^T\in \Re^{n}$ is the joint rate vector. Also, $J_{Te}\in\Re^{3\times n}$ and $J_{Re}\in\Re^{3\times n}$ are the translational and rotational Jacobian matrices, respectively. The skew-symmetric matrix $\left[ \mathbf{r}^{\times} \right] \in \Re^{3\times 3}$ is defined for any vector $\mathbf{r} =[r_1,r_2,r_3]^T \in \Re^{3}$ as:
\begin{equation} 
    \left[ \mathbf{r}^{\times} \right] = \begin{bmatrix}
        0 & -r_3 & r_2\\
        r_3 & 0 &  -r_1\\
        -r_2 & r_1 & 0
    \end{bmatrix}.
\end{equation}
The translational and rotational Jacobian matrices are defined as:
\begin{equation}
    J_{Te} = \left[\hat{\mathbf{k}}_1 \times (\mathbf{r}_e - \mathbf{p}_1),
    \ldots, \hat{\mathbf{k}}_n \times (\mathbf{r}_e-\mathbf{p}_n) \right],
    \label{Eq_6}
\end{equation}
\begin{equation}
    J_{Re} = \left[\hat{\mathbf{k}}_1, \hat{\mathbf{k}}_2,
    \ldots, \hat{\mathbf{k}}_n\right],
    \label{Eq_7}
\end{equation}
where ${\hat{\bf k}}_i \in \Re^3$ is the unit vector along the rotation axis of joint $i$, and ${\bf p}_i \in \Re^3$ is the position vector of joint $i$ in the inertial frame. 

Equation (\ref{Eq_d1}) can be rewritten as:
\begin{equation}
    \begin{bmatrix}
\dot{\mathbf{r}}_e \\
\boldsymbol{\omega}_e 
\end{bmatrix} = J_b \begin{bmatrix}
    \dot{\mathbf{r}}_b\\
  \boldsymbol{\omega}_b
\end{bmatrix} + J_m \dot{\mathbf{q}},
\label{Eq_8}
\end{equation}
where $J_b \in \Re^{6\times6}$ is the base-spacecraft Jacobian, and $J_m \in \Re^{6\times n}$ is the manipulator Jacobian. 

The equations of motion for the SMS are derived using the Lagrangian approach, where the Lagrange function is defined as $L = T-V$, with $T$ and $V$ being the kinetic and potential energy of the system, respectively. In a microgravity environment, the potential energy caused by gravity is negligible, and for a small free-floating SMS structure, the Lagrangian equals the kinetic energy \cite{schaub2003analytical}:
\begin{equation}
    L = T = \frac{1}{2}\left[\left(m_b \dot{\mathbf{r}}_b^T \dot{\mathbf{r}}_b +\boldsymbol{\omega}_b^T I_b\boldsymbol{\omega}_b\right)+\sum_{i=1}^{n}\left(m_i \dot{\mathbf{r}}_i^T \dot{\mathbf{r}}_i + \boldsymbol{\omega}_i^T I_i \boldsymbol{\omega}_i\right)\right].
    \label{Eq_LT}
\end{equation}
where ${\bf r}_i \in \Re^3$ and $\bm\omega_i \in \Re^3$ are the position and angular velocity of the link-$i$'s center of mass in the inertial frame, respectively. The kinetic energy can be expressed in a compact form as:
\begin{equation}
    T = \frac{1}{2} \begin{bmatrix}
\dot{\mathbf{r}}_b \\
\boldsymbol{\omega}_b\\
\dot{\mathbf{q}}
\end{bmatrix}^T \begin{bmatrix}
    H_v & H_{v\omega} & H_{vm}\\
    H_{v\omega}^T & H_{\omega} & H_{\omega m}\\
    H_{vm}^T & H_{\omega m}^T & H_m
\end{bmatrix} \begin{bmatrix}
    \dot{\mathbf{r}}_b\\
  \boldsymbol{\omega}_b\\
    \dot{\mathbf{q}}
\end{bmatrix}.
\label{Eq_T}
\end{equation}
The matrices, $H_v\in \Re^{3\times3}$, $H_{v\omega}\in \Re^{3\times3}$, $H_{\omega}\in \Re^{3\times3}$, $H_{vm}\in \Re^{3\times n}$, $H_{\omega m}\in \Re^{3\times n}$, and $H_m\in \Re^{n\times n}$, are defined as:
\begin{equation}
    H_v = m_c \mathbb{I}_3,
    \label{Eq_14}
\end{equation}
\begin{equation}
    H_{v\omega} = m_c (\mathbf{r}_{cb}^{\times})^T,
\end{equation}
\begin{equation}
    H_{\omega} = I_b +\sum_{i=1}^{n} (I_i + m_i (\mathbf{r}_{ib}^{\times})^T (\mathbf{r}_{ib}^{\times})) ,
\end{equation}
\begin{equation}
    H_{vm} =\sum_{i=1}^{n} m_i J_{Ti},
\end{equation}
\begin{equation}
   H_{\omega m} = \sum_{i=1}^{n} (I_i J_{Ri} + m_i (\mathbf{r}_{ib}^{\times}) J_{Ti}),
\end{equation}
\begin{equation}
    H_m = \sum_{i=1}^{n} (J_{Ri}^T I_i J_{Ri} + m_i J_{Ti}^T J_{Ti}),
\end{equation}
where $I_i$ and $m_i$ are the moment of inertia matrix and mass of 
the $i$-th link (for $i=1, \dots, n$),  $m_c = m_b + \sum_{i=1}^{n} m_i$ is the total mass of the system,  $\mathbf{r}_c = 
\left(\mathbf{r}_b m_b +\sum_{i=1}^{n}\mathbf{r}_i m_i\right)/m_c$ is the position vector of the whole system's center of mass, and ${\bf r}_{ib} = {\bf r}_i -  {\bf r}_b$. The translational and rotational Jacobian matrices, $J_{Ti} \in \Re^{3\times n}$ and $J_{Ri} \in \Re^{3\times n}$, are defined as:
\begin{equation}
     J_{Ti} = \left[\hat{\mathbf{k}}_1 \times (\mathbf{r}_i - \mathbf{p}_1),...., \hat{\mathbf{k}}_i \times (\mathbf{r}_i-\mathbf{p}_i), \mathbb{O}_{3\times n-i} \right]\,\,\,\forall (1\leq i\leq n),
\end{equation}
\begin{equation}
     J_{Ri} = \left[\hat{\mathbf{k}}_1, \hat{\mathbf{k}}_2,...., \hat{\mathbf{k}}_i, \mathbb{O}_{3\times n-i}\right]\,\,\,\forall (1\leq i\leq n).
\end{equation}
The generalized coordinates used to describe the SMS are chosen as $\boldsymbol{\Phi} =[\mathbf{r}_b, \bm{\epsilon}_{\upsilon}, \mathbf{q}]^T \in \Re^{6+n}$, representing the combination of the base spacecraft's linear ($\mathbf{r}_b$) and angular ($\mathbf{\epsilon}$) positions, and the manipulator joint angles $\mathbf{q}$. The Lagrange equation can be written as:
\begin{equation}
    \frac{d}{dt}\left(\frac{\delta L}{\delta \dot{\mathbf{\Phi}}}\right) -\frac{\delta L}{\delta \mathbf{\Phi}} = \bm{\tau},
    \label{Eq_lagrange_1}
\end{equation}
where $\bm{\tau}\in\Re^{6+n}$ is the generalized force vector, which includes the driving force/torque on the base body during orientation control and the manipulator joint torque. Substituting Eqs. (\ref{Eq_LT}) and (\ref{Eq_T}) into Eq. (\ref{Eq_lagrange_1}) yields the following equations of motion (corresponding to $n+6$ scalar equations) for the free-floating SMS:
\begin{equation}
    H(\boldsymbol{\Phi}) \ddot{\boldsymbol{\Phi}} + C(\dot{\boldsymbol{\Phi}},\boldsymbol{\Phi}) \dot{\boldsymbol{\Phi}} = \bm{\tau},
    \label{Eq_EOM}
\end{equation}
where $H(\boldsymbol{\Phi}) \in \Re^{(6+n)\times(6+n)}$ is the generalized inertial matrix, and $C(\dot{\boldsymbol{\Phi}},\boldsymbol{\Phi}) \in \Re^{(6+n)\times(6+n)}$ is the Coriolis and centrifugal matrix of the system. This matrix represents velocity-dependent forces and includes the dynamic coupling terms between the base and the manipulator(s), especially in free-floating systems. Note that the first six elements of $\bm{\tau}$, which correspond to control inputs for the translational and rotational motion of the base, are zero in a free-floating system.

\subsection{Dynamic Coupling in Free-Floating SMS}
Considering that external forces and torques acting on the SMS are negligible, the free-floating SMS operates based on the principle of momentum conservation. Assuming the initial values of 
angular and linear momentum are zero, the linear and angular momentum
equations (${\bf h}_l\in\Re^{3}$ and ${\bf h}_a\in\Re^{3}$) with respect to the whole system's center of mass can be written as \cite{xu1993measure}:
\begin{equation}
    \mathbf{h}_l = m_b \dot{\mathbf{r}}_b + \sum_{i=1}^{n} m_i \dot{\mathbf{r}}_i =
    \mathbf{0},
    \label{Eq_P}
\end{equation}
\begin{equation}
    \mathbf{h}_a = (I_b \boldsymbol{\omega}_b + \mathbf{r}_b \times m_b \dot{\mathbf{r}}_b) + \sum_{i=1}^{n} (I_i \boldsymbol{\omega}_i + \mathbf{r}_i \times m_i \dot{\mathbf{r}}_i) =
    \mathbf 0.
    \label{Eq_L}
\end{equation}
Equations (\ref{Eq_P}) and (\ref{Eq_L}) can be expressed in matrix form as:
\begin{equation}
    \begin{bmatrix}
        {{\bf h}_l}\\
        {{\bf h}_a}
    \end{bmatrix} = \begin{bmatrix}
        H_v & H_{v\omega} \\
        H_{v\omega}^T & H_{\omega}
    \end{bmatrix} \begin{bmatrix}
        \dot{\mathbf{r}}_b\\
        \boldsymbol{\omega}_b
    \end{bmatrix} + \begin{bmatrix}
        H_{vm}\\
        H_{\omega m}
    \end{bmatrix} \dot{\mathbf{q}} = 
    \mathbf 0.
    \label{Eq_DC_1}
\end{equation}

Note that the coefficient matrix (composed of $H_v$, $H_{v\omega}$, and $H_\omega$), representing the mass and inertia of the system, is always invertible. Therefore, a direct relationship between the base velocities and the joint rates can be found as:
\begin{equation}
    \begin{bmatrix}
        \dot{\mathbf{r}}_b\\
        \boldsymbol{\omega}_b
    \end{bmatrix} =  
    C_{bm} \dot{\mathbf{q}},
    \label{Eq_DC_2}
\end{equation}
where $C_{bm} \in \Re^{6\times n}$ is the joint-to-base dynamic coupling matrix, defined as:
\begin{equation}
    C_{bm} = - \begin{bmatrix}
        H_v & H_{v\omega} \\
        H_{v\omega}^T & H_{\omega}
    \end{bmatrix}^{-1} \begin{bmatrix}
        H_{vm}\\
        H_{\omega m}
    \end{bmatrix}.
\end{equation} 
Inserting Eq. (\ref{Eq_DC_2}) into Eq. (\ref{Eq_8}) yields:
\begin{equation}
     \begin{bmatrix}
\dot{\mathbf{r}}_e \\
\boldsymbol{\omega}_e 
\end{bmatrix} = (J_b C_{bm} + J_m)\dot{\mathbf{q}}\ =J^*\dot{\mathbf{q}},
\label{Eq_DC_GJM}
\end{equation}
and the joint rates can be found as:
  \begin{equation}
      \dot{\mathbf{q}} = {J^*}^{\dagger}
      \begin{bmatrix}
\dot{\mathbf{r}}_e \\
\boldsymbol{\omega}_e  
\end{bmatrix}.
\label{Eq_DC_3}
  \end{equation}
{In Eq. (\ref{Eq_DC_3}), the matrix $J^*$ is known as the generalized Jacobian matrix, which relates joint velocities to end-effector velocities while eliminating the base velocity using momentum conservation \cite{wilde2018equations}. 
Finally, inserting Eq. (\ref{Eq_DC_3}) into Eq. (\ref{Eq_DC_2}) yields:
\begin{equation}
\begin{bmatrix}
    \dot{\mathbf{r}}_b\\
        \boldsymbol{\omega}_b
    \end{bmatrix} = 
C_{be} \begin{bmatrix}
\dot{\mathbf{r}}_e \\
\boldsymbol{\omega}_e  
\end{bmatrix},
\label{Eq_DC_4}
\end{equation}
where $C_{be} = C_{bm} {J^*}^{\dagger} \in \Re^{6\times 6}$ is the end-to-base dynamic coupling matrix. From Eqs. (\ref{Eq_DC_2}), (\ref{Eq_DC_3}), and (\ref{Eq_DC_4}), it is evident that joint rate variation affects the end-effector trajectory as well as the satellite motion.      

\section{Optimal Trajectory Design and Trajectory Tracking Control}
This section introduces a new approach for dynamic coupling-informed optimal trajectory design. The proposed algorithm analyzes and leverages 
dynamic coupling effects in a free-floating SMS to enable end-effector operations without active base control.

\subsection{Designing an Optimal Trajectory Informed by Dynamic Coupling}
In most existing literature, dynamic coupling effects in free-floating space manipulators are treated as disturbances, and various onboard control strategies have been proposed to suppress or compensate for the resulting base motion. In contrast, this work presents a trajectory optimization approach that does not aim to minimize dynamic coupling
. Instead, it strategically leverages dynamic coupling by incorporating it into the cost function. The proposed cost formulation rewards dynamic coupling when it constructively supports the end-effector's motion in the desired direction and penalizes it when it hinders task performance.

To understand and quantify the dynamic coupling between the base and the manipulator in a free-floating SMS, this work performs an SVD of the dynamic coupling matrix $C_{bm}$. The SVD is a powerful matrix factorization technique that expresses the dynamic coupling matrix $C_{bm} \in \Re^{6\times n}$ as:
\begin{equation}
    C_{bm} = {\bf{U}\bm{\Sigma}\bf{V}}^T,
    \label{eq:SVD}
\end{equation}
where $\mathbf{U}\in \Re^{6\times 6}$ and $\mathbf{V}\in \Re^{n\times n}$ are orthogonal matrices, and $\bm{\Sigma} \in \Re^{6\times n}$ is a diagonal matrix containing the singular values
. Note that the {$i$-th} column vector
(${\bf{v}}_i$) of $\bf{V}$ represents the principal direction in the joint space, that is, the characteristic joint motion direction that serves as input to the system. The $i$-th diagonal entry
(${\sigma}_i$) of $\bm{\Sigma}$ quantifies 
the amplification or strength of coupling associated with each of the joint direction. The $i$-th column vector (${\bf{u}}_i$) of $\bf{U}$, on the other hand, represents the corresponding direction of motion induced in the base. Thus, the SVD reveals how specific joint motion patterns contribute to base motion and to what extent. 

The first singular vector pair (${\bf{v}}_1$, ${\bf{u}}_1$), associated with $\sigma_1$, provides the dominant coupling direction information. Moving the joints in the combination given by ${\bf{v}}_1$ will induce the largest base motion, specifically in the direction ${\bf{u}}_1$. In contrast, 
the last singular pair (${\bf{v}}_n$, ${\bf{u}}_n$) corresponds to the least effective way to move the base (or potentially a near-null-space motion if $\sigma_n \approx 0$). Thus, if the joint velocity vector is aligned with ${\bf{v}}_n$, the base barely moves. 
 
To define the dynamic coupling-informed cost function, two metrics are introduced to quantify the quality of dynamic coupling: the normalized Shannon entropy ($H_{norm}$) of the singular value spectrum and the coupling assistance metric ($\cos(\theta_{a})$). 

First, the normalized entropy of the singular value spectrum 
measures how evenly the coupling strengths are distributed across all directions. It 
gives a quantitative measure of how uniformly or unevenly the manipulator joint motions affect the base motion. The normalized Shannon entropy is calculated as \cite{strydom2021svd}:
\begin{equation}
H_{norm} = {-}\frac{
\sum_{i=1}^{n} s_i \log(s_i)}{\log(n)}, 
\label{eqn_SVD_2}
\end{equation}
where $s_i ={\sigma_i}/{\sum_{j=1}^{n}\sigma_j}$ and $H_{norm}\in [0,1]$. 
A value of $H_{norm} =0$ indicates highly directional coupling, where most of the energy is concentrated in a single mode. That is, one singular value dominates the spectrum. In contrast, $H_{norm} =1$ corresponds to uniformly distributed coupling, where 
energy is spread equally across all modes, and the singular values are comparable in magnitude.

Second, the coupling assistance metric $\cos(\theta_{a})$ is calculated based on the dynamic coupling direction and the desired end-effector movement direction. This metric evaluates whether the dynamic coupling-informed joint space direction aids the end-effector in reaching its desired position. As previously discussed, $\mathbf{v}_1$ represents the dominant 
dynamic coupling direction, whereas $\mathbf{v}_n$ represents the weakest. This work employs the joint velocity of the manipulator using a combination of the dynamic coupling directions as:
\begin{equation} 
    \dot{\mathbf{q}}_{DC} =\sum_{i=1}^{n} \alpha_i \mathbf{v}_i, 
    \label{eqn_SVD_3}
\end{equation}
where $\alpha_i \in [-1,1]$ are scalar scaling factors. Let the final desired end-effector position be $\mathbf{r}_d \in \Re^3$. By substituting Eq. (\ref{eqn_SVD_3}) into Eq. (\ref{Eq_DC_GJM}), one obtains $\dot{\mathbf{r}}_{e_{DC}}$, which represents the end-effector motion induced by the combination of dynamic coupling directions. To assess whether this motion aligns with the direction toward the desired position, the coupling assistance metric is defined as: 
\begin{equation}
    \cos(\theta_a) = \frac{\hat{\mathbf{d}}^T {\mathbf{r}}_{e_{DC}}}{||\hat{\mathbf{d}}|| \ ||\dot{\mathbf{r}}_{e_{DC}}||},
    \label{eqn_SVD_4}
\end{equation}
where $\hat{\mathbf{d}}\in \Re^3$ is the unit vector from the current end-effector position to the desired 
position, computed as $\hat{\mathbf{d}} =\frac{\mathbf{r}_d - \mathbf{r}_e}{|| \mathbf{r}_d - \mathbf{r}_e||}$. A value of $\cos(\theta_a) = 1$ indicates that the resulting motion fully assists the end-effector in reaching its goal while a value of  $\cos(\theta_a) = 0$ implies a neutral effect, offering no direct assistance. Conversely, $\cos(\theta_a) = -1$ indicates that the motion directly opposes the desired 
motion. 

Based on the two metrics, $H_{norm}$ and $\cos(\theta_a)$, the joint trajectory optimization problem is formulated to find the set of scaling factors ($\alpha_i$) that minimizes the following cost function:
\begin{equation}
    J = \sum_{k=1}^{N} (\tilde{C}(k) - (1-H_{norm}(k)))^2,
    \label{eq:optimal_cost}
\end{equation}
subject to
\begin{align}
    & |{q}_{i}(k)| \leq {q}_{{i}_{max}},\hspace{0.1cm} i=1,2, {\cdots, n}
    , \label{eq:const1} \\
    & |\dot{{q}}_{i}(k)| \leq \dot{{q}}_{{i}_{max}},\hspace{0.1cm} i=1,2,{\cdots, n}
    ,\label{eq:const2} \\
    & |\ddot{{q}}_{i}(k)| \leq \ddot{{q}}_{{i}_{max}}, \hspace{0.1cm} i=1,2,{\cdots, n}
    , \label{eq:const3} \\
    & ||\mathbf{d}_i(k) - \mathbf{d}_j(k)||\geq {d}_{safe}, \forall i<j, \label{eq:const4} \\
    & |\mathbf{f}_{i}^{(1)}(k)| > \frac{l}{2},\hspace{0.1cm} \forall i\geq 2, \label{eq:const5}
    \\
    & |\mathbf{f}_{i}^{(2)}(k)| > \frac{w}{2},\hspace{0.1cm} \forall i\geq 2, \label{eq:const6}
    \\
    & |\mathbf{f}_{i}^{(3)}(k)| > \frac{h}{2},\hspace{0.1cm} \forall i\geq 2, \label{eq:const7}
    \\
    & || \mathbf{r}_d - \mathbf{r}_e(k)|| \leq {r_{th}}
    , \hspace{0.1cm} \forall k \in [0.9T,T], \label{eq:const8}
\end{align}
where $\tilde{C}$ is defined as $\frac{1}{{\kappa}} \log(1+\exp({\kappa}\cos(\theta_{a})))$, ${q}_{{i}_{max}}$, $\dot{{q}}_{{i}_{max}}$, and $\ddot{{q}}_{{i}_{max}}$ denote the maximum joint angle, velocity, and acceleration, respectively. Equation (\ref{eq:const4}) enforces that any two adjacent points, as shown in Fig
ure \ref{fig:SMS_collision_diagram}
, must maintain a safe distance to prevent collisions between the links. 
\begin{figure}[htbp]
	\centering\includegraphics[width=5.5in]{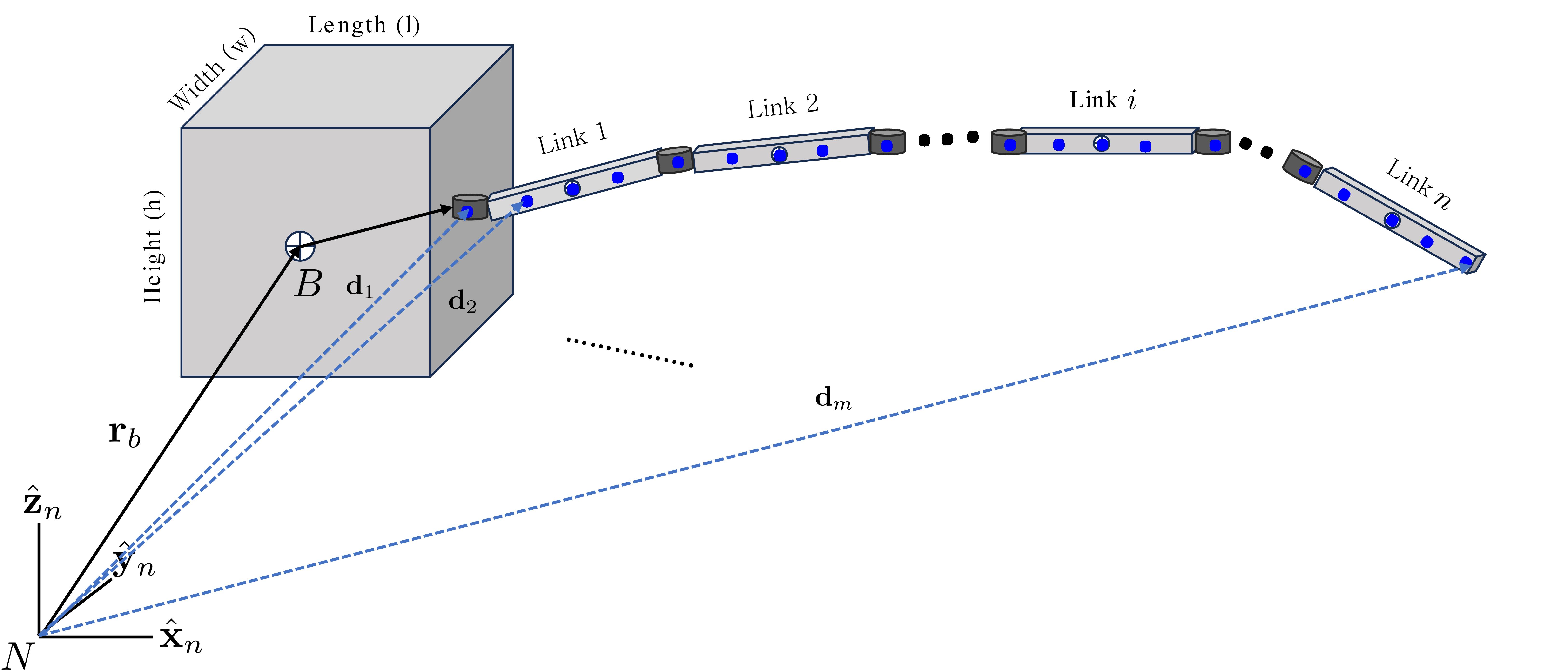}
	\caption{Schematic of the Free-Floating SMS Highlighting the Collision Check Points.}
	\label{fig:SMS_collision_diagram}
\end{figure}
Similarly, to avoid collision between the manipulator and the satellite base, each manipulator point must satisfy the constraints given in Eqs. (\ref{eq:const5}) - (\ref{eq:const7}). In 
these equations, $\mathbf{f}_i = \mathbf{d}_i - \mathbf{r}_b$, and $l$, $w$, and $h$ are the length, width, and height of the satellite base, respectively. The final constraint, given in Eq. (\ref{eq:const8}), states that during the last 10\% of the simulation time, the norm of the position error between the actual and desired end-effector positions must remain within a specified bound.

In summary, the objective is to optimize the weights $\alpha_i$ in Eq. (\ref{eqn_SVD_3}) to compute joint velocities that effectively leverage dynamic coupling information for optimal trajectory design. The resulting joint velocities 
at the $k$-th time step are computed as follows:
\begin{equation} 
        \dot{\mathbf{q}}_{opt}{(k)} = {\sum_{i=1}^{n} \alpha_{i_{opt}(k)} \mathbf{v}_i(k)}.
    \label{eq:optimal_joint_Vel}
\end{equation}

The base motion is computed as $\dot{\mathbf{x}}_{b_{opt}}{(k)}= C_{bm} \dot{\mathbf{q}}_{opt}{(k)}$, and the end-effector position is obtained by substituting $\dot{\mathbf{q}}_{opt}$ into Eq. (\ref{Eq_DC_GJM}). Similarly, $\mathbf{q}_{opt}$ and $\ddot{\bf q}_{opt}$ can be obtained by integrating and differentiating $\dot{\mathbf{q}}_{opt}$, respectively. 

\subsection{Sliding Mode Control Design for Trajectory Tracking}
After generating the optimal trajectory, the next step is to validate whether the onboard control system can accurately follow the trajectory and guide the end-effector to the desired position without inducing unwanted motion in the base. In this work, an SMC is designed for trajectory tracking. The sliding surface is defined as:
\begin{equation}
    \mathbf{s} = \dot{\mathbf{q}}_{e} + {\Gamma}
    \mathbf{q}_{e}, 
    \label{eqn:sliding_surface}
\end{equation}
where $\mathbf{q}_{e} = \mathbf{q} - \mathbf{q}_{opt}${, $\dot{\mathbf{q}}_{e} = \dot{\mathbf{q}} - \dot{\mathbf{q}}_{opt}$,} and $\Gamma \in 
{\Re}^{3 \times 3}$ is a symmetric positive definite gain matrix
. The joint control torque $\bm{\tau}_q \in \Re^{n\times1}$ is designed as: 
\begin{equation}
    \bm{\tau}_q = H_q (\ddot{\mathbf{q}}_{opt} - \Gamma \dot{\mathbf{q}}_e) + C_q {+ \bm{\tau}_{q_{sw}}},
    \label{eqn:control_torque}
\end{equation}
where $H_q \in \Re^{n\times n}$ and $C_q \in \Re^{n\times 1}$ are expressed as:
\begin{equation}
    H_q = H_m -  \begin{bmatrix}
        H_{vm}\\
        H_{\omega m}
    \end{bmatrix}^T\begin{bmatrix}
        H_v & H_{v\omega} \\
        H_{v\omega}^T & H_{\omega}
    \end{bmatrix}^{-1} \begin{bmatrix}
        H_{vm}\\
        H_{\omega m}
    \end{bmatrix},
\end{equation}
\begin{equation}
    C_q = C_m - \begin{bmatrix}
        H_{vm}\\
        H_{\omega m}
    \end{bmatrix}^T\begin{bmatrix}
        H_v & H_{v\omega} \\
        H_{v\omega}^T & H_{\omega}
    \end{bmatrix}^{-1} C_b.
\end{equation}
Here, the $C_b \in \Re^{6\times1}$ and $C_m\in \Re^{n\times1}$ 
are sub-matrices that form the  Coriolis and centrifugal matrix, defined as:
\begin{equation}
    C(\dot{\boldsymbol{\Phi}},\boldsymbol{\Phi}) \dot{\boldsymbol{\Phi}} = \begin{bmatrix}
    C_b(\dot{\boldsymbol{\Phi}},\boldsymbol{\Phi})\\
    C_m(\dot{\boldsymbol{\Phi}},\boldsymbol{\Phi})
\end{bmatrix}.
\end{equation}

The switching torque is defined as $\bm{\tau}_{q_{sw}}=- K_s  \text{sat}\left(\frac{{\bf s}}{\lambda}\right)$, where $\lambda$ is a positive scalar value, and $K_s$ is a positive definite matrix. Note that the saturation function is defined as:
\begin{equation}
     \text{sat}\left( \frac{\mathbf{s}_i}{\lambda} \right) =
\begin{cases}
1 & \text{if } \frac{\mathbf{s}_i}{\lambda} > 1 \\
\frac{\mathbf{s}_i}{\lambda} & \text{if } -1 \leq \frac{\mathbf{s}_i}{\lambda} \leq 1 \\
-1 & \text{if } \frac{\mathbf{s}_i}{\lambda} < -1
\end{cases}
\quad \text{for } i = 1, 2,\cdots, n
.
\end{equation}
Using Lyapunov stability analysis provided in the Appendix, it can be shown that the closed-loop system is asymptotically stable during the reaching phase and maintains sliding mode stability once on the sliding surface.
\section{Results and Discussion}
This study illustrates the planar motion of a free-floating SMS equipped with a single robotic arm consisting of three links to validate the performance of the proposed approach. Figure \ref{fig:Scenario1} illustrates the SMS configuration and {the} desired end-effector locations used in this study. 
\begin{figure}[htbp
]
    \centering
    \includegraphics[width=0.6\textwidth]{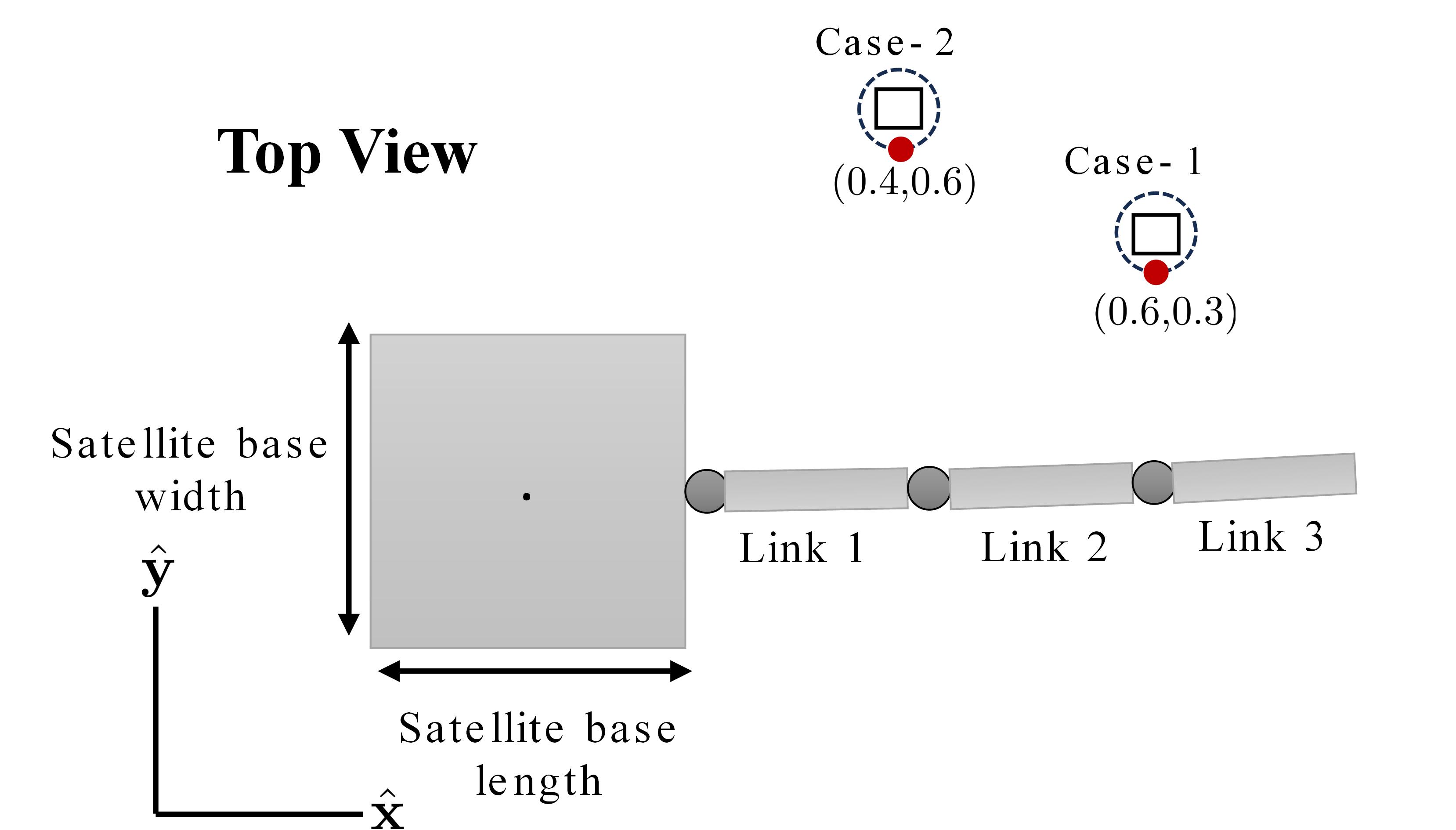}
    \caption{Illustration of the SMS Configuration}
\label{fig:Scenario1}
\end{figure}
The rectangular body surrounded by the dotted circle represents the intended target on which the SMS gripper executes tasks{,} such as grasping, maintenance, and handling operations. To avoid any potential collision{s} between the manipulator gripper and the object during these operations, the end-effector is required to reach a designated red point located within {an} approximately 
10 mm perimeter around the object. All the simulation parameters are summarized in Tables \ref{tab:Simulation_param} and \ref{tab:Simulation_param2}. 
\begin{table}[htbp]
\centering
\caption{Simulation Parameters for SMS}
\renewcommand{\arraystretch}{1.3}
\begin{tabular}{>{\centering\arraybackslash}p{2.7cm}|>{\centering\arraybackslash}p{6cm}|c|c|c}
\hline
\textbf{
Parameters} & \textbf{Base} & \textbf{Link 1} & \textbf{Link 2} & \textbf{Link 3} \\
\hline
Mass (kg) & 31.015 & 0.569 & 0.569 & 0.569 \\
$I_{xx}$ (kg$\cdot$m$^2$) & 1.1594 & 0.0001 & 0.0001 & 0.0001 \\
$I_{yy}$ (kg$\cdot$m$^2$) & 1.1594 & 0.0043 & 0.0043 & 0.0043  \\
$I_{zz}$ (kg$\cdot$m$^2$) &  1.1129 & 0.0043 & 0.0043 & 0.0043 \\
l{ength} (m)& 0.464 & 0.3 & 0.3 & 0.3 \\
w{idth} (m)& 0.464 & 0.03 & 0.03 & 0.03 \\
h{eight} (m)& 0.483 & 0.03 & 0.03 & 0.03 \\
\multirow{2}{*}{\shortstack{{Initial values in}\\ {the }position{-level}}} & ${^N{\bf r}_b=[-0.0356, -0.0006, 0]^T}$ {(}m{)} 
& \multirow{2}{*}{1 {(deg)}}
& \multirow{2}{*}{1 {(deg)}}
& \multirow{2}{*}{1 {(deg)}}
\\
 & $\bm{\epsilon} =[0,0,0,1]^T$  & & & \\
\multirow{2}{*}{\shortstack{{Initial values in}\\ {the }velocity{-level}}} &${^N{\bf \dot{r}}_b=[0, 0, 0]^T}$ {(}m/s{)}
& \multirow{2}{*}{0 (deg/s)} & \multirow{2}{*}{0 (deg/s)} & \multirow{2}{*}{0 (deg/s)}\\
 & $\bm{\omega}_b =[0,0,0]^T$ {(}deg/s{)} & & & \\ \hline
\end{tabular}
\label{tab:Simulation_param}
\end{table}

\begin{table}[h
]
\centering
\caption{Simulation Parameters for the Optimization and Controller }
\renewcommand{\arraystretch}{1.3}
\begin{tabular}{c|c|c|c|c|c}
\hline
\multicolumn{2}{c|}{\textbf{Parameters}} & \multicolumn{4}{c}{\textbf{Values }} \\
\hline
\multirow{5}{*}{Optimization} & $\mathbf{q}_{max}$ 
& \multicolumn{4}{c}{$[81,162,162]^T$ {(deg)}}\\
& $\dot{\mathbf{q}}_{max}$ 
& \multicolumn{4}{c}{$[22.92,22.92,22.92]^T$ {(deg/s)}} \\
& $\ddot{\mathbf{q}}_{max}$ 
& \multicolumn{4}{c}{$[28.65,28.65,28.65]^T$ {(deg/$s^2$)}}\\
& ${d}_{safe}$ 
& \multicolumn{4}{c}{$0.01$ {(m)}}\\
& ${r}_{th}$ 
& \multicolumn{4}{c}{$0.02$ {(m)}} \\
\hline
\multirow{4}{*}{Controller} &$K_s$ & \multicolumn{4}{c}{diag([0.001
{,}0.001
{,}0.001])} \\
& $\Gamma$ & \multicolumn{4}{c}{diag([10
{,}10
{,}10])} \\
& $\lambda$ & \multicolumn{4}{c}{0.02} \\
& $\bm \tau_{q_{max}}$ 
& \multicolumn{4}{c}{$[3.5,1.5,1.5]^T$ {(Nm)}}  \\
\hline
\end{tabular}
\label{tab:Simulation_param2}
\end{table}
In Table \ref{tab:Simulation_param}, $I_{xx}$, $I_{yy}${,} and $I_{zz}$ denote the principal moments of inertia about the base's {$x$}-, {$y$}-, and {$z$}-axes, respectively, expressed in their respective body frames. The inertial frame is assumed to be attached to the center of mass of the entire system. All three joints are revolute and rotate about the $z$-axis, represented by the unit vector $\hat{\bf{k}} =[0,0,1]^T$. Since setting all joint angles to zero would result in a singularity, the initial configuration described in Table \ref{tab:Simulation_param} is selected to avoid this issue. The manipulator parameters listed in Table \ref{tab:Simulation_param} are based on a hardware testbed on the air-bearing system 
currently under development for validating this simulation study 
. 

Starting from the initial configuration, two cases are considered, each with a different desired end-effector position{,} as 
shown in Figure \ref{fig:Scenario1}. For both cases, a dynamic coupling-informed optimal trajectory is first generated using Eq. (\ref{eq:optimal_joint_Vel}). The designed sliding mode controller then ensures that the end-effector follows the optimal trajectory to reach the desired position.

\subsection{Case-1: Short-Range Manipulation Task } 
This case demonstrates end-effector trajectory generation and control for a target position located close to the initial end-effector location, as shown in Figure \ref{fig:Scenario1}, highlighting precise manipulation with minimal disturbance to the base. Figure \ref{fig:opt_traj_Case_1} illustrates the optimal trajectory generated using the dynamic coupling-informed approach. 
\begin{figure}[htbp]
    \centering
    \begin{subfigure}[b]{0.49\textwidth}
        \centering
        \includegraphics[width=\textwidth]{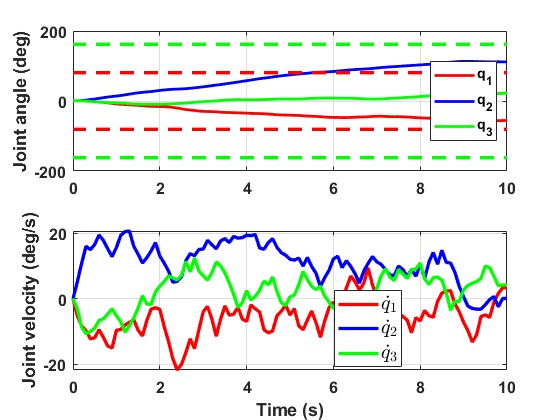}
        \caption{Manipulator Joint Motion}
        \label{fig:case1_joint}
    \end{subfigure}
    \hfill
    \begin{subfigure}[b]{0.49\textwidth}
        \centering
        \includegraphics[width=\textwidth]{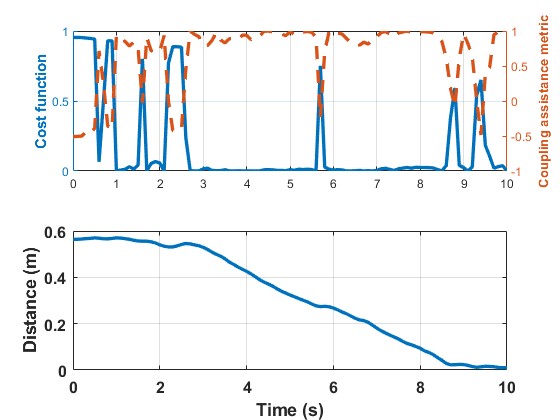}
        \caption{Cost Function, Coupling Assistance Metric, and End-Effector Relative Position Error}
        \label{fig:case1_cost}
    \end{subfigure}
    \vspace{0.03cm}    
    \begin{subfigure}[b]{0.49\textwidth}
        \centering
        \includegraphics[width=\textwidth]{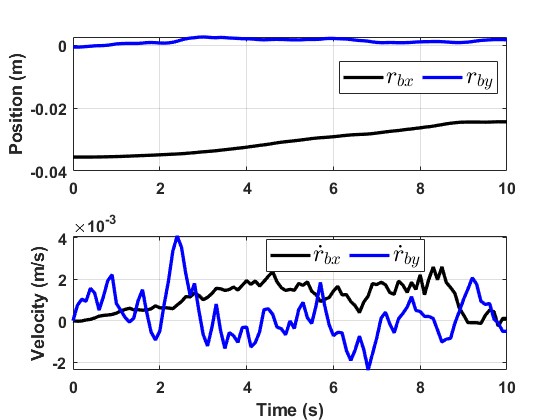}
        \caption{Satellite Base Linear Motion}
        \label{fig:case1_linear}
    \end{subfigure}
    \hfill
    \begin{subfigure}[b]{0.49\textwidth}
        \centering
        \includegraphics[width=\textwidth]{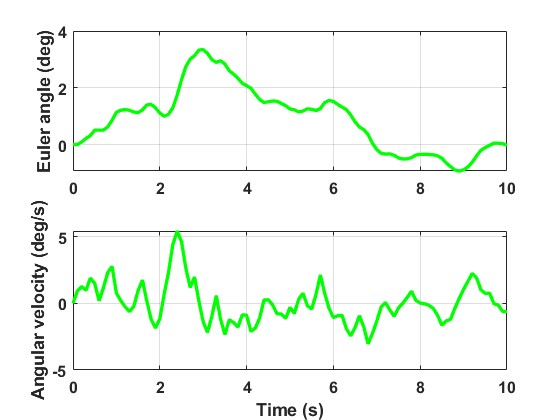}
        \caption{Satellite Base Angular Motion}
        \label{fig:case1_angular}
    \end{subfigure}
    \caption{Time History of the Generated Optimal Trajectory for the Free-Floating SMS (Case-1)}
    \label{fig:opt_traj_Case_1}
\end{figure}
Figure \ref{fig:case1_joint} shows that both {the} optimal joint angles and joint velocities remain within their respective limits (shown 
{by} the dotted lines) throughout the motion.
From Figures \ref{fig:case1_linear} and \ref{fig:case1_angular}, it is evident that the base motion{,} propagated by the optimal joint trajectories{,} experiences only minor motion along the computed trajectory. Additionally, 
Figure \ref{fig:case1_cost} shows the evolution of the dynamic coupling-informed cost function and the coupling assistance metric. It can be observed that a lower coupling assistance metric corresponds to a higher cost function, indicating increased dynamic interaction, which in turn affects joint velocities and base motion. The frequent oscillation of the coupling assistance metric between positive and negative values suggests that whenever the manipulator’s joint motion is not perfectly aligned with the dynamic preferences of the base, 
alternating assistance and resistance in the coupled dynamics 
{are} induced. This dynamic interaction causes the induced base velocities to fluctuate between small positive and negative values, thereby maintaining a bounded and minimal net base displacement over time. A steady reduction in the end-effector’s relative position error is also observed in Figure \ref{fig:case1_cost}, confirming effective convergence toward the desired end-effector position. 

The performance of the controller in tracking the optimal trajectory is presented in Figure \ref{fig:ControlErr_case_1}. As shown in Figure \ref{fig:case1_error}, the joint angle and joint velocity errors remain small and bounded, indicating accurate tracking. Figure \ref{fig:case1_torque} illustrates the consistent decrease in the end-effector’s position error, along with the corresponding joint torques required to achieve the motion. 
\begin{figure}[htbp]
   \centering
    \begin{subfigure}[b]{0.49\textwidth}
        \centering
        \includegraphics[width=\textwidth]{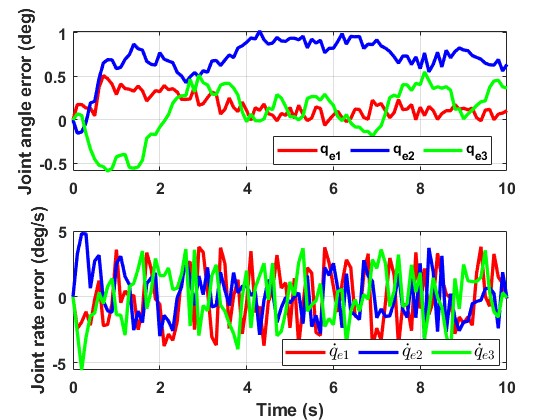}
        \caption{
        Joint Angle and Joint Rate Errors}
        \label{fig:case1_error}
    \end{subfigure}
    \begin{subfigure}[b]{0.49\textwidth}
        \centering
       \includegraphics[width=\textwidth]{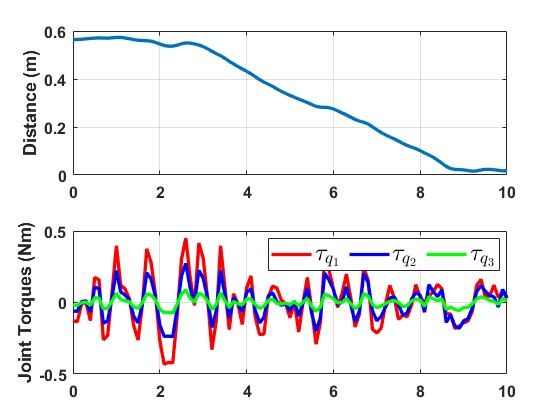}
        \caption{End-Effector Relative Position Error and Required Joint Control Torques} 
        \label{fig:case1_torque}
    \end{subfigure}
    \caption{Time History of the Controlled Trajectory for the Free-Floating SMS (Case-1)}
    \label{fig:ControlErr_case_1}
\end{figure}


Overall, the results of Case{-}1 confirm that the developed trajectory enables the end-effector to reach its nearby target while minimizing base disturbance and avoiding internal collisions. Moreover, the controller ensures accurate joint-level tracking of the optimal trajectory.


\subsection{Case-2: Long-Range Manipulation Task} 
This case illustrates the end-effector trajectory generation and control for a desired position {that is} significantly distant from the initial end-effector location, as shown in Figure \ref{fig:Scenario1}, requiring substantial arm reconfiguration and dynamic base interaction. Figure \ref{fig:opt_traj_Case_2} illustrates the generated optimal joint trajectories
{, along with} the corresponding states and cost value. 
\begin{figure}[htbp]
    \centering
    \begin{subfigure}[b]{0.49\textwidth}
        \centering
        \includegraphics[width=\textwidth]{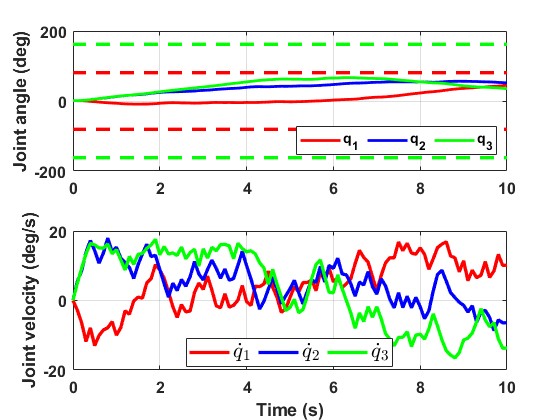}
        \caption{Manipulator Joint Motion}
        \label{fig:case2_joint}
    \end{subfigure}
    \hfill
    \begin{subfigure}[b]{0.49\textwidth}
        \centering
        \includegraphics[width=\textwidth]{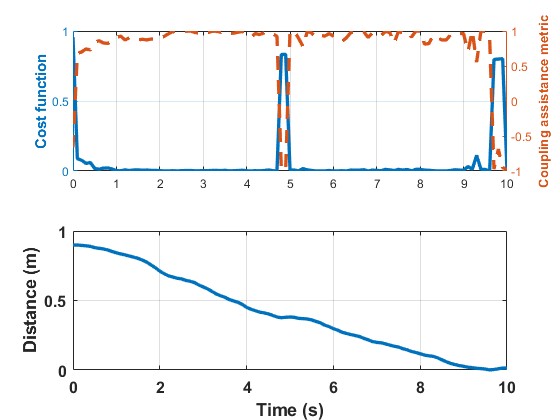}
        \caption{Cost Function, Coupling Assistance Metric, and End-Effector Relative Position Error}
        \label{fig:case2_cost}
    \end{subfigure}
    \vspace{0.03cm}    
    \begin{subfigure}[b]{0.49\textwidth}
        \centering
        \includegraphics[width=\textwidth]{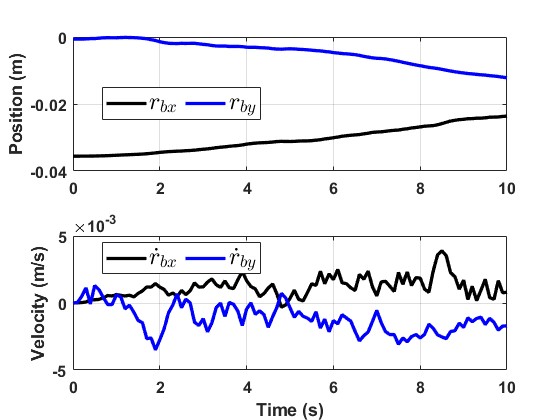}
        \caption{Satellite Base Linear Motion}
        \label{fig:case2_linear}
    \end{subfigure}
    \hfill
    \begin{subfigure}[b]{0.49\textwidth}
        \centering
        \includegraphics[width=\textwidth]{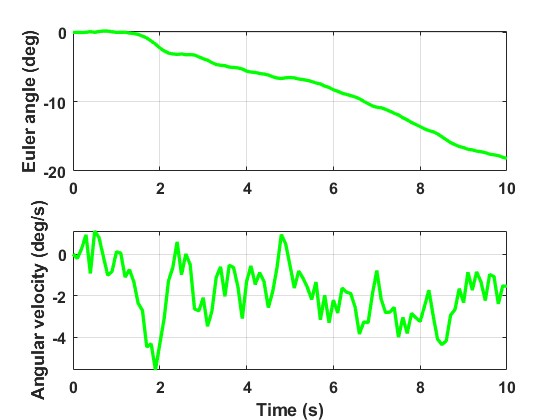}
        \caption{Satellite Base Angular Motion}
        \label{fig:case2_angular}
    \end{subfigure}
    \caption{Time History of the Generated Optimal Trajectory for the Free-Floating SMS (Case-2)}
    \label{fig:opt_traj_Case_2}
\end{figure}
Unlike the local manipulation task in Case-1, this case involves a distant target end-effector position, resulting in significant base motion, as evident from Figures \ref{fig:case2_linear} and \ref{fig:case2_angular}. Joint angles gradually evolve toward their respective final configurations with
{out} constraint violations, similar to Case-1, as shown in Figure \ref{fig:case2_joint}. However, joint velocities demonstrate more oscillations and quadrant switching, indicating the need for directional adjustments during execution{,} as shown in Figure \ref{fig:case2_joint}. In Figure \ref{fig:case2_cost}, the dynamic coupling-informed cost value shows sharp peaks at specific intervals (around 0–1 s, 4–5 s, and near 10 s), corresponding to moments 
when the coupling assistance metric drops, signaling phases of lower alignment between joint motion and favorable base dynamics. During these intervals, joint-level motions likely create disturbances in the base, requiring compensatory joint velocity direction change{,} as can be seen in Figure \ref{fig:case2_joint}. The bottom plot in Figure \ref{fig:case2_cost} illustrates that the end-effector position error decreases smoothly, confirming that despite the higher dynamic complexity, the optimizer successfully ensures convergence to the desired end-effector position. 

The controller's performance in tracking the optimal trajectory for Case-2 is shown in Figure \ref{fig:ControlErr_case_2}. 
\begin{figure}[htbp]
   \centering
    \begin{subfigure}[b]{0.49\textwidth}
        \centering
        \includegraphics[width=\textwidth]{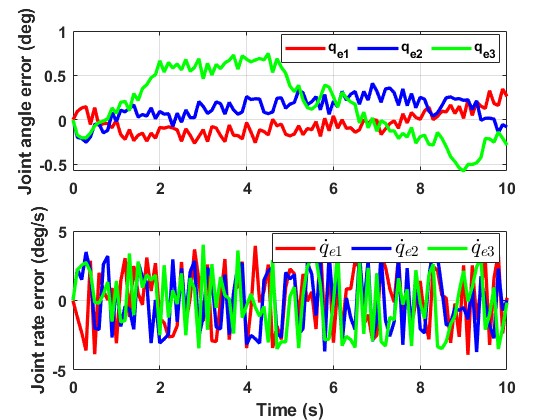}
        \caption{
        Joint Angle and Joint Rate Errors}
        \label{fig:case2_error}
    \end{subfigure}
    \begin{subfigure}[b]{0.49\textwidth}
        \centering
       \includegraphics[width=\textwidth]{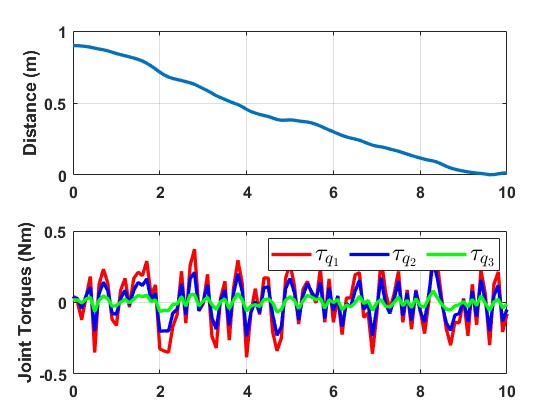}
        \caption{End-Effector Relative Position Error and Required Joint Control Torques} 
        \label{fig:case2_torque}
    \end{subfigure}
    \caption{Time History of the Controlled Trajectory for the Free-Floating SMS (Case-2)}
    \label{fig:ControlErr_case_2}
\end{figure}
As seen in Figure \ref{fig:case2_error}, the joint angle and velocity errors remain bounded, demonstrating stable tracking despite {the} increased dynamic complexity. Figure \ref{fig:case2_torque} shows a steady reduction in the end-effector position error, along with the joint torques required to execute the trajectory. In contrast to Case-1, 
Case-2 shows that the induced base motion is significantly larger, yet {it} effectively aids the end-effector in reaching the distant target under the designed sliding mode controller.

Overall, the Case-2 results show that for distant manipulation tasks, base motion becomes an active contributor to the trajectory execution. The optimizer effectively balances manipulator effort and base dynamics using the coupling-aware cost function. The induced base motion, though larger than {in} Case-1, remains bounded, and the joint states evolve within limits, validating the robustness of the proposed trajectory generation approach for dynamic and large-scale maneuvers. In this case 
{as well}, the controller ensures accurate joint-level tracking of the optimal trajectory.



\section{Conclusion}
Space manipulator systems (SMSs) play a crucial role in on-orbit servicing (}OOS), enabling tasks like repair, refueling, and debris removal that significantly extend satellite lifespans and reduce the high costs of launching replacements. Traditionally, dynamic coupling in free-floating systems has been regarded as a disturbance to be suppressed through active base control, reinforcing the preference for free-flying systems. However, in practical scenarios{,} such as close-proximity operations or actuator failures, free-floating conditions are often unavoidable, necessitating control strategies that function without direct base actuation. 
While prior research has predominantly focused on minimizing this coupling, often overlooking its potential advantages, this work investigates
how dynamic coupling can 
be leveraged while operating SMSs. By extracting key features from the dynamic coupling matrices, this work 
proposed a framework for a dynamic coupling-informed trajectory optimization algorithm. Simulation results demonstrate that explicitly incorporating dynamic coupling into the trajectory planning process leads to more informed and efficient system behavior, enhancing operational safety and autonomy in future OOS missions.
As future work, this work will be validated through hardware experiments and further developed based on a higher degree-of-freedom robotic arm operated in 3-dimensional space.

\section{Acknowledgment}
This material is based upon work supported by the Air Force Office of Scientific Research under award number FA9550-24-1-0600.

\section{Appendix}\label{sec:app}
This section provides a stability proof for the proposed sliding mode controller. First, it is shown that the control law defined in Eq. (\ref{eqn:control_torque}) ensures that all trajectories converge to the sliding manifold. Let the Lyapunov function be defined by a radially unbounded, positive definite function:
\begin{equation}
    V_r = \frac{1}{2} \mathbf{s}^T H_q \mathbf{s}{,}
    \label{eqn:V_r_1}
\end{equation}
where $H_q$ is a symmetric positive definite matrix that varies smoothly with the system state. 
Taking the time derivative of both sides of Eq. (\ref{eqn:V_r_1}) yields:
\begin{equation}
    \dot{V}_r = \frac{1}{2} \mathbf{s}^T \dot{H}_q \mathbf{s} + \mathbf{s}^T H_q \dot{\mathbf{s}}{.}
    \label{eqn:V_r_2}
\end{equation}
Note that in the Lyapunov analysis, the derivative of ${H}_q$ is neglected under the assumption that $\dot{H}_q$ is bounded and relatively small compared to the dominant sliding term. Inserting the control input from Eq. (\ref{eqn:control_torque}) into Eq. (\ref{eqn:V_r_2}) results in
{:}
\begin{equation}
    \dot{V}_r \leq - K_s \mathbf{s}^T  \text{sat}\left(\frac{\mathbf{s}}{\lambda}\right){.}
\end{equation}
The time derivative of the Lyapunov function is negative semi-definite. Hence, the sliding surface 
is attractive, and the closed-loop system is 
{Lyapunov} stable
. 

Once the system states reach the sliding surface $\mathbf{s} =\bf 0
$, the trajectories converge to the origin in a stable manner. Let a Lyapunov function be defined in terms of joint errors as:
\begin{equation}
    V_s = \frac{1}{2} \mathbf{q}_e^T \mathbf{q}_e.
    \label{eqn:V_s_1}
\end{equation}

Taking the time derivative of both sides of Eq. (\ref{eqn:V_s_1}) yields:
\begin{equation}
    \dot{V}_s = \mathbf{q}_e^T \dot{\mathbf{q}}_e.
    \label{eqn:V_s_2}
\end{equation}

On the sliding surface, $\mathbf{s} = \bf 0$, which implies $\dot{\mathbf{q}}_e = -\Gamma
\mathbf{q}_e$ from Eq. (\ref{eqn:sliding_surface}). Substituting into Eq. (\ref{eqn:V_s_2}) gives:
\begin{equation}
    \dot{V}_s = -2{\Gamma}
    V_s. 
\end{equation}
Since $\Gamma$ is a symmetric positive definite matrix, {$\dot{V}_s$ is negative definite, and the system is exponentially stable.
\bibliographystyle{AAS_publication}   
\bibliography{references}   

\end{document}